\documentclass[letterpaper, 10 pt, conference]{ieeeconf}  % Comment this line out if you need a4paper

\IEEEoverridecommandlockouts                              % This command is only needed if 
\usepackage{graphics} % for pdf, bitmapped graphics files
\usepackage{subfigure}
\usepackage{amsmath}
\usepackage{amssymb}
\usepackage{array}
\usepackage{multirow}
\usepackage[pdftex]{graphicx}
\usepackage{eqparbox}
\usepackage{float}
\usepackage{booktabs}
\usepackage{tabu}
\usepackage{url}
\usepackage{dsfont}
\usepackage{color, colortbl}
\usepackage{caption}
\usepackage{tabularx}
\usepackage{booktabs, cellspace, multirow, tabularx}
\newcolumntype{C}[1]{>{\centering\arraybackslash}p{#1}}
\title{\LARGE \bf
Unlock the Power of Unlabeled Data in Language Driving Model
}

\author{
Chaoqun Wang$^{1,2}$,\thanks{$^{1}$ Sun Yat-sen University.}\thanks{$^{2}$ The Chinese University of Hong Kong, Shenzhen. \{chaoqunwang, jieyang5\}@link.cuhk.edu.cn}
Jie Yang$^{1,2}$,
Xiaobin Hong$^{3}$, \thanks{$^{3}$ NanJing University. xiaobinhong@smail.nju.edu.cn}
and
Ruimao Zhang$^{1\dag}$\thanks{\dag Corresponding author. ruimao.zhang@ieee.org}\\
}

\begin{document}

\maketitle
\thispagestyle{empty}
\pagestyle{empty}

%%%%%%%%%%%%%%%%%%%%%%%%%%%%%%%%%%%%%%%%%%%%%%%%%%%%%%%%%%%%%%%%%%%%%%%%%%%%%%%%
\begin{abstract}
Recent Vision-based Large Language Models~(VisionLLMs) for autonomous driving have seen rapid advancements. However, such promotion is extremely dependent on large-scale high-quality annotated data, which is costly and labor-intensive.
To address this issue, we propose unlocking the value of abundant yet unlabeled data to improve the language-driving model in a semi-supervised learning manner.
Specifically, we first introduce a series of template-based prompts to extract scene information, generating questions that create pseudo-answers for the unlabeled data based on a model trained with limited labeled data. Next, we propose a Self-Consistency Refinement method to improve the quality of these pseudo-annotations, which are later used for further training.
By utilizing a pre-trained VisionLLM (e.g., InternVL), we build a strong Language Driving Model (LDM) for driving scene question-answering, outperforming previous state-of-the-art methods.
Extensive experiments on the DriveLM benchmark show that our approach performs well with just 5\% labeled data, achieving competitive performance against models trained with full datasets.
In particular, our LDM achieves 44.85\% performance with limited labeled data, increasing to 54.27\% when using unlabeled data, while models trained with full datasets reach 60.68\% on the DriveLM benchmark.

\end{abstract}

\section{Introduction}
Recent advancements in Vision Large Language Models for autonomous driving have made significant progress~\cite{tian2024drivevlm, wang2023drivemlm, mao2023gpt}. These models follow a common process, which perceives the driving scene and makes driving decisions by asking the corresponding questions, making driving scene VQA a critical task.
However, as application scenarios continue to diversify, with increasingly stringent precision requirements, the need for large volumes of annotated data to improve model performance becomes more pressing, meanwhile, large-scale annotating is both labor-intensive and time-consuming. This naturally raises the following question:
\textit{\textbf{How to train the Language Driving Model with readily collectible unlabeled data?}}

A straightforward and intuitive approach is to follow a semi-supervised learning paradigm, which aims to generate pseudo labels for unlabeled data using a model trained on a few labeled seed data. Nevertheless, for autonomous driving with complex VQA tasks, 
this strategy presents several challenges: \textbf{\textit{1) Diverse questions}}: How can we involve diverse questions that allow the model to infer meaningful VQA annotations? \textbf{\textit{2) High-quality answer}}: The quality of the predicted pseudo label is heavily coupled to the performance of the trained LDM and is crucial for sequential training.

\begin{figure}[tp]
    \centering
    \includegraphics[width=0.99\linewidth]{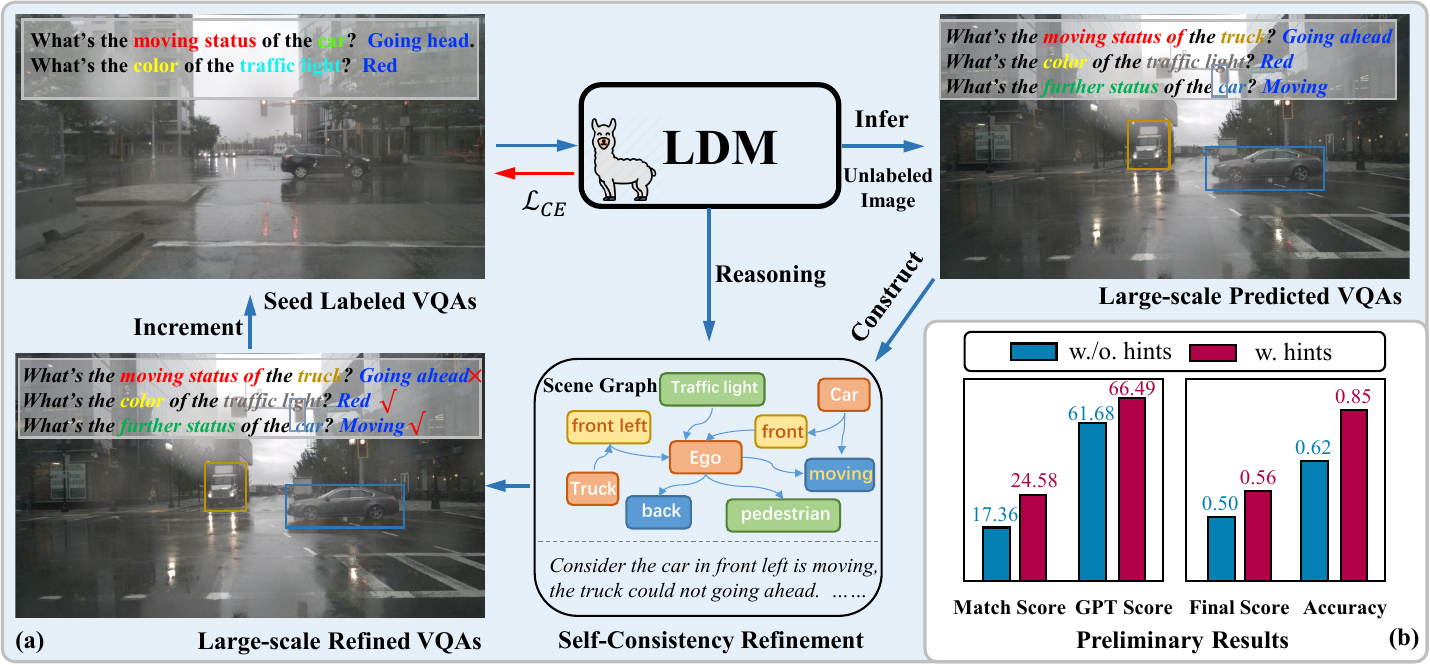}
    \vspace{-1mm}
    \caption{(a): The proposed semi-supervised Language Driving Mode~(LDM) traning pipeline. We first fine-tune an LDM in a supervised manner with a few labeled data. Then, given a fine-tuned LDM and unlabeled data, we generate VQAs via templated-based prompts and refine them by the proposed Self-Consistency Refinement~(SCR). The refined VQAs are used for subsequent model training. (b): Preliminary experiment on DriveLM benchmark, inferring the fine-tuned LDM with hints could obtain better performance.}
    \label{fig:intro}
\vspace{-5mm}
\end{figure}

To obtain high-quality and diverse pseudo annotations, we attempt to utilize the inherent rules in the driving scenes of the physical world.
Empirical practice illustrates that in language-driving annotation, an image of a scene is often paired with multiple QAs, which display multiple objects related to the ego vehicle and their interactions.
Therefore, another intuition arises as to \textbf{\textit{whether these connections and interactions in scenes could help LDMs better understand the objects.}} To verify the positive effect of the \textit{``hints''} for the object understanding, we conduct preliminary experiments as shown in Fig.~\ref{fig:intro}-(b). 
Instead of just predicting object states using question prompts, we extracted cues from the scene and added them to the prompts for better predictions. Finally, we found that by using the prompts with hints, LDM could obtain more accurate predictions.

Based on the above observations, we propose a novel iterative training pipeline to unlock the potential of readily available unlabeled data in a semi-supervised learning manner as shown in Fig.~\ref{fig:intro}-(a).
\textbf{\textit{Firstly}}, we introduce a series of template-based prompts to extract the scene information by excavating the status of the objects related to safe driving, which could provide diverse questions to infer the LDM given multi-view images for a driving scene, which is trained with seed labeled data.
\textbf{\textit{Secondly}}, we construct the scene graph to adequately represent important objects in the scene and the relationships between them based on the generated multiple QA pairs of a driving scene, which provides important clues to predict their future state.
\textbf{\textit{Thirdly}}, we retrieve many hints from the scene graph for each QA pair and re-ask the question with the hints for a better response. Furthermore, we propose Self-Consistency Refinement (SCR) that considers various hints to achieve the refinement of pseudo-labels with arbitrary precision. In SCR, two different prompts for the same question constitute an important accordance for the trustworthiness computation of the sample. As shown in Fig.~\ref{fig:intro}-(a), given a fine-tuned LDM and large-scale unlabeled data, our proposed SCR could produce high-quality VQA pairs for the subsequent model training. 
In such a manner, we could constantly promote the LDM ability by interactively increasing the unlabeled data beginning
with seed annotated data.

To demonstrate the effectiveness of our proposal, we conduct experiments on the DriveLM benchmark and deliver an iterative training setting. 
First, we meticulously design the prompts to unlock the potential of large models in driving scenes and gain state-of-the-art performances based on the off-the-shelf pre-trained VisionLLM~(e.g. InternVL2).
Then, to show the effectiveness of the proposed interactive training pipeline based on the semi-supervised learning paradigm, we train the model with seed ~(5\%) annotations and gain huge promotion with iterative incremental large-scale~(20\% and 75\% for 2 iterations) unlabeled data.
Specifically, our model trained on the full annotation achieves a \textbf{60.68}\% final score, which is comparable with the past competition\footnote{\url{https://opendrivelab.com/challenge2024/#driving_with_language}} winner~(60.02\% final score). 
Meanwhile, for the semi-supervised setting, our model gains 44.85\% with 5\% labels. By employing our proposed interactive training pipeline with 95\% unlabeled images, our proposal earns a 54.27\% final score and +\textbf{9.42}\% promotion.

The main contributions are summarized as follows:
\begin{itemize}
    \item As far as we know, this is the first work to unlock the potential of unconstrained unlabeled visual data to improve the Language Drive Model in a semi-supervised training paradigm.
    \item We propose a novel self-consistency refinement method to improve the quality of generated pseudo-labels via the graph-based hints, paving a new path to scale up training data.
    \item We conduct extensive experiments to demonstrate the effectiveness of our proposal and gain remarkable promotions compared to the baseline model. 
\end{itemize}

\section{Related Work}

\subsection{Large Driving Model}
Vision-based Large Language Models (VisionLLMs) seamlessly integrate Natural Language Processing (NLP) with Computer Vision (CV), which is instrumental in advancing applications like scene perception, planning, question answering, and content generation in autonomous driving.
For example, leveraging the visual-text alignment capabilities of vision-language pre-trained models like CLIP, several prior works~\cite{cheng2023language, wei2024bev, tang2024bev} have demonstrated success in detecting open-vocabulary objects in autonomous driving scenes. Additionally, some approaches~\cite{wu2023language, liao2024vlm2scene} integrate cross-modal features into a prompt reasoning framework to predict 3D objects, while recent studies~\cite{chahe2024leveraging, zheng2024large, peng2024lc, ding2023hilm} directly use VisionLLMs to perceive complex driving scenarios and behaviors, such as identifying long-tail risk objects or detecting abnormal driving patterns.
In recent years, there has been a surge in research focused on language-guided decision-making~\cite{wen2023dilu, xu2024drivegpt4, tian2024drivevlm, wang2023drivemlm} and trajectory planning~\cite{mao2023gpt, pan2024vlp, sharan2023llm}, and shows strong cognitive and reasoning abilities
To further enhance the alignment of vision and language features, many works have incorporated VisionLLMs with diffusion models to construct world models grounded in language, as seen in projects like DriveDreamer~\cite{wang2023drivedreamer, zhao2024drivedreamer2}, GALA~\cite{hu2023gaia}, DrivingDiffusion~\cite{li2023drivingdiffusion}, and DriveGenVLM~\cite{fu2024drivegenvlm}.
Unlike previous approaches, our work aims to unlock the potential of unlabeled driving scene images.
% eliminating the need for costly and time-consuming manual annotation. 
Through a semi-supervised training strategy, we seek to enhance the capabilities of VisionLLMs by leveraging these abundant, unlabeled datasets, advancing the field of autonomous driving further.

\subsection{Semi-Supervised Learning}
Traditional model training follows supervised learning has yielded significant success but depends heavily on labeled data, which is expensive and time-consuming to obtain~\cite{dridi2021supervised,el2021supervised}. Semi-supervised learning~(SSL) addresses this limitation by leveraging both labeled and unlabeled data, making it particularly valuable for industries where extensive labeled datasets are difficult or costly to acquire~\cite{canedo2019facial,allabadi2023semi}. SSL improves model performance while reducing the need for manual labeling~\cite{chi2019semi,mey2022improved}.
Recent advances in SSL were driven by improvements in deep learning architectures~\cite{synnaeve2019end}, optimization techniques~\cite{chapelle2008optimization}, and data augmentation strategies~\cite{frommknecht2022augmentation}. SSL approaches typically fall into two categories: pseudo-labeling~\cite{zhu2023alternative} and consistency regularization~\cite{lin2021semifed}, both of which effectively harness labeled and unlabeled data during training.
Large Language Models (LLMs), pre-trained on vast amounts of text data, have recently been used to generate pseudo-labels for SSL training, offering a zero-shot label inference mechanism. This LLM-based pseudo-labeling~\cite{pham2021meta,chen2020big} has proven to be an efficient method for SSL, as LLMs can generate human-like annotations and transfer knowledge through distillation~\cite{wang2021want,ding2022gpt,kim2022ask,su2022selective}.
To further enhance pseudo-label quality, prior methods relied on human or expert models like GPT to rank generated labels~\cite{ding2022gpt,gu2024minillm,xu2023baize}. In contrast, our approach refines pseudo-labels using a self-consistency mechanism, eliminating the need for human labeling or powerful expert models.
\section{Method}
\label{sec:method}

In this section, we first present the overall framework of the proposed Language Driving Model in Sec.~\ref{sec:framework}, and then detail introduces the iterative training pipeline in Sec.~\ref{sec:training_pipeline} that iterative utilizes the unlabeled data.
Finally, we introduce the proposed scene graph construction process in Sec.~\ref{sec:question}, and the self-consistency refinement in Sec.~\ref{sec:self_consistency}, which generate the questions and pseudo labels to extract the scene information from images, and refine the predictions for the subsequent training respectively.

\begin{figure}[t]
\vspace{2mm}
    \centering
    \includegraphics[width=0.9\linewidth]{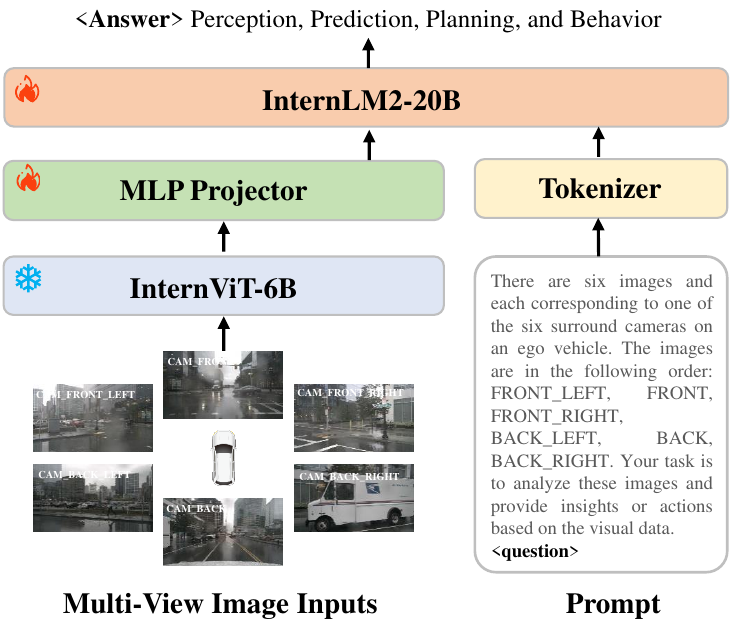}
    % \vspace{-1mm}
    \caption{Overall framework of Language Driving Model. Given multi-view images, we feed them into a shared InternViT-6B. Then, the language tokens from the prompt and the projected vision tokens are fed into the InternLM2-20B model to generate the responses.}
    \label{fig:model}
\vspace{-5mm}
\end{figure}

\subsection{Network Architecture.}
\label{sec:framework}
As shown in Fig.~\ref{fig:model}, the overall proposed Language Driving Model~(LDM) is based on an off-the-shelf VisionLLM~(e.g. InternVL2), which gains state-of-the-art performances in the scenes understanding and question answering. 
Given multi-view images of a driving scene, we first watermark the camera view information on the corresponding image for a clearer statement. Then, to understand the multi-images with surrounding views, we feed them into a shared Vision Transformer~(e.g. InternViT-6B) to extract the vision tokens. To better align with the language inputs, the vision tokens are transformed by an MLP projector composed of attention and multi-layer perceptron components. 
For the language prompts, we adopt an InternLM2 tokenizer to obtain the corresponding language tokens.
The vision and language tokens are incorporated together and fed into a Large Language Model~(e.g. InternLM2-20B) to generate the final response.

\subsection{Iterative Training Pipeline.}
\label{sec:training_pipeline}
In our proposed iterative language driving model training pipeline, we aim to unlock the potential of the readily available unconstrained unlabeled data in a semi-supervised training paradigm. In which we propose a graph-based Vision-Question-Answer~(VQA) generation method to extract the scene information and a Self-Consistency Refinement~(SCR) method to improve the quality of the generated pseudo labels for the subsequent training.
\begin{figure*}[htbp]
    \centering
    \includegraphics[width=0.95\linewidth]{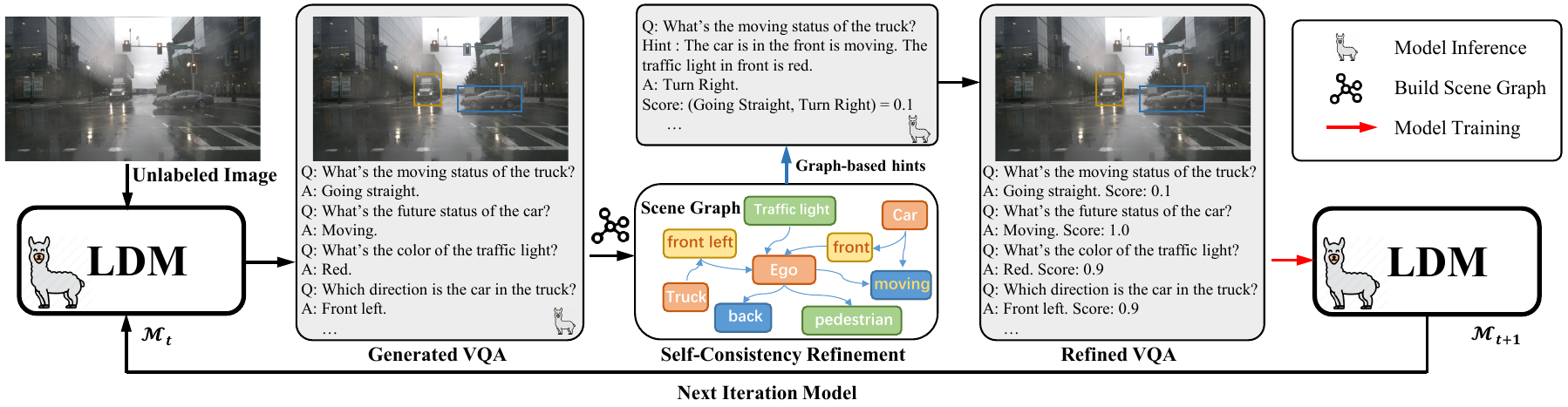}
    \vspace{-1mm}
    \caption{Overall iterative training pipeline. Given an unlabeled image, we utilize the fine-tuned LDM $\mathcal{M}_t$ to generate the multiple VQAs and build the scene graph from the predictions. Then, by extracting graph-based hints, the question could be re-asked. By calculating the distance between the two predictions, we could obtain the reliable score for each prediction which is utilized as the sample-wise balance weight to train the subsequent model $\mathcal{M}_{t+1}$ in the next iteration.}
    \label{fig:train}
\vspace{-5mm}
\end{figure*}
Initially, our approach is based on the premise that we have a pre-trained VisionLLM $\mathcal{M}$ and a limited-scale dataset $\mathcal{D}_0 = \{(v_i, q_i, a_i)\}^{n_0}_{i=1}$ including $n_0$
labeled VQA pairs, denoting the corresponding image, question, and answer.
We could train a LDM in a supervised manner:
\begin{align}
    \pi^{*}_{\theta} = \mathop{\arg\min}_{\pi_{\theta}} \mathbb{E}_{\{v_i, q_i,a_i\} \sim \mathcal{D}_0} \mathcal{L}(\mathcal{M}(v_i, q_i, \pi_{\theta}), a_i),
\end{align}
where $\pi_{\theta}$ is the parameter of the LDM and $\mathcal{L}$ is the objective function. We could obtain a fine-tuned LDM $\mathcal{M}_0$.

Then, we could iteratively train the LDM by introducing incremental unlabeled data. As shown in Fig.~\ref{fig:train}, given a fine-tuned model $\mathcal{M}_t$ at the $t$-th iteration, our proposal aims to generate high-quality pseudo labels from readily available images to train a subsequent model $\mathcal{M}_{t+1}$.

Specifically, given a large-scale images set $\{v_i\}^{n_t}_{i=1}$ without any annotation, we first recognize the elements of the scene via the finetuned $\mathcal{M}_t$ that are relevant to the ego vehicle driving and generate many questions to extract scene information based on manually set template-based prompts, which detail introduce in Sec.~\ref{sec:question}. Then, the $\mathcal{M}_t$ is employed to generate the dataset $\mathcal{D}_t = \{(v_i, q_i, a_i)\}^{n_t}_{i=1}$, where the $a_i$ is the $i$-th predicted pseudo labels. 

Then, based on the multiple QA pairs of a sample, we refine the predicted pseudo labels via our proposed Self-Consistency Refinement and obtain a high-quality scored VQAs to train the subsequent LDM $\mathcal{M}_{t+1}$. In practice, based on the multiple QA pairs of a sample, we could build the scene graph, where the nodes are the recognized elements. For each node, we could define the attributions such as visual description, meaning, moving status, and edges such as directions, distance, and influence. Further, we retrieve hints from the scene graph and re-ask the question based on the hints. If the re-ask prediction is similar to the original prediction, we consider it more credible, and vice versa. The more detailed introduction is referred to Sec.~\ref{sec:self_consistency}.

Finally, based on the incremental refined dataset  $\mathcal{D}_t = \{(v_i, q_i, a_i, s_i)\}^{n_t}_{i=1}$, we could train the subsequent model as follows:
\begin{align}
\begin{split}
\pi^{*}_{\theta} = \mathop{\arg\min}_{\pi_{\theta}} \mathbb{E}_{\{v_i, q_i, a_i, s_i\} \sim \tilde{\mathcal{D}_t}} \mathcal{L}(\mathcal{M}_t(v_i, q_i, \pi_{\theta}), a_i, s_i),
\end{split}
\end{align}
where $\tilde{\mathcal{D}_t} = \{\mathcal{D}_0, \cdots, \mathcal{D}_t\}$ indicates mixing the $t$ datasets, $\pi^{*}_{\theta}$ is the optimized parameter of $\mathcal{M}_{t+1}$, and $0 \leq s_i \leq 1$ is the credible scores obtained by the proposed SCR method. For the $\mathcal{D}_0$ with annotation, we set $s_i=1$.
In such a manner, we could iteratively utilize more unlabeled data and continuously promote the performance of the LDM.

\subsection{Scene Graph Construction}
\label{sec:question}
For a given sample, we could construct a scene graph $G=(V, E)$, where $V=\{o_1, o_2, \cdots, o_m\}$ are the $m$ selected objects in the current scene and $E=\{e_{1,1}, \cdots e_{i,j}, \cdots e_{m,m} \}$ are the connections within node pairs. Specifically, for each node $o_i$, we use $\mathcal{A}_i$ to indicate the attributions such as color, visual description, status, etc. Edge $e_{i,j}$ denotes whether node $o_i$ is connected to node $o_j$, and the edge feature $\mathcal{E}_{i,j}$ means the relationships within node-pair, including the direction, moving direction, influences, etc.
We introduce the node selection and attribution/edge generation in the following, which are achieved by a series of template-based prompts.
% It's worth noting that we achieve the above process by a series of template-based prompts.

\textbf{Node selection.} In DriveLM, the model chooses the node by the following prompt: \textit{``What are the important objects in the current scene? Those objects will be considered for future reasoning and driving decisions."}. With such a prompt, the LDM will respond to multiple nodes $\mathcal{O}=\{o_1, o_2, ..., o_m\}$, in which $o_i = <c_i, CAM, x, y>$, where $c_i$ is the identifier, $CAM$ is the camera view, and $(x,y)$ is the object location. To relieve the inherent challenges of the VisionLLM in predict a precision coordinate, we employ a detection model such as YOLO~\cite{reis2023real}, to detect all the object coordinates relevant to the driving, including car, truck, pedestrian, traffic light, traffic signs, etc. Based on the detected boxes incorporating the aforementioned prompt, the fine-tunes LDM could select 3$\sim$5 important objects as nodes.

\textbf{Attribution generation.} For each node, we define its attributions via the templated-based class-relevant prompt. First, we obtain the object's visual description with the following prompts: \textit{``What is the visual description of the object $o_i$?"}. Then, for the moveable objects, we mainly obtain their moving status with the following prompt:  \textit{``What is the observed status of the object $o_i$?"}, \textit{"What is the moving status of the object $o_i$?"}, and \textit{``What is the future status of the object $o_i$?"}, etc. For the traffic sign/light, we mainly focus on its meaning by the following prompt: \textit{``What is the meaning of the object $o_i$?"}. Based on the above prompts, the fine-tuned LDM could generate responses as pseudo labels.

\textbf{Edge generation.} For each node pair, we build connections according to the distances within them, which is similar to the attributions but with different prompts.
Specifically, for two close objects, we mainly focus on their moving directions and interaction effect with the following prompts: \textit{``Which direction is $o_i$ from $o_j$?"}, \textit{``Based on $o_i$, what's the action of $o_j$?"}, and \textit{``What actions taken by the $o_i$ can lead to a collision with $o_j$?"}, etc.

\begin{table*}[h!]
\centering
\caption{The performance comparison on the test set. $^{*}$ indicates the reproduced method. VQA indicates the annotated data and V is the image without label. Acc. and C.G are the short of Accuracy and chatGPT. $\text{LDM}_i$ indicates the results of the $i$-th iteration of our proposal. The best results are shown in \textbf{bold}.}
\vspace{-2mm}
\resizebox{1.0\linewidth}{!}{
\begin{tabular}{cccccccccccccc}
\toprule
Method & VQA & V & Acc. & C.G & Bleu\_1 & Bleu\_2 & Bleu\_3 & Bleu\_4 & ROUGE\_L & CIDEr & Match & Final Score \\
\midrule
ImagineLab~\cite{li2024driving}  & 100\%  & 0\% & 0.73 & 65.25 & \textbf{0.78} & \textbf{0.72} & \textbf{0.66} & \textbf{0.61} & \textbf{0.74} & \textbf{0.21} & 47.65 &  0.6002 \\
MTMM~\cite{anonymous2024mmad}  & 100\%  & 0\% & 0.75 & \textbf{65.58} & 0.76 & 0.70 & 0.64 & 0.59 & 0.74 & 0.18 & 45.00 &0.5987 \\
LDM &  100\% & 0\% & \textbf{0.77} & \textbf{65.58} & 0.73 & 0.68 & 0.62 & 0.57 &	0.73 & 0.17 & \textbf{49.21} &	\textbf{0.6068} \\

\hline
ImagineLab$^{*}$ & 5\% & 0\% & 0.63 & 56.09 & 0.17 & 0.13 & 0.10 & 0.08 & 0.25 & 0.01 &36.82 & 0.4479 \\
$\text{LDM}_0$ & 5\% & 0\% & 0.63 & 56.35 & 0.17 & 0.13 & 0.10 & 0.08 & 0.25 & 0.01 & 36.61 & 0.4485 \\
$\text{LDM}_1$ & 5\% & 20\% & 0.64 & 56.28 & 0.56 & 0.50 & 0.45 & 0.39 & 0.39 & 0.08 & \textbf{40.11} & 0.4915 \\
$\text{LDM}_2$ & 5\% & 95\% & \textbf{0.69} & \textbf{63.06} & \textbf{0.67} & \textbf{0.59} & \textbf{0.56} & \textbf{0.50} &\textbf{ 0.70} & \textbf{0.11} & 35.22 & \textbf{0.5427}\tiny(+9.42\%) \\
\bottomrule

\end{tabular}}
\label{tab:main_test}
\vspace{-3mm}
\end{table*}

\begin{table*}[htbp]
\centering
\caption{The performance comparison on the validation set.}
\vspace{-2mm}
\resizebox{1.0\linewidth}{!}{
\begin{tabular}{cccccccccccccc}
\toprule
Method & VQA & V & Acc. & C.G. & Bleu\_1 & Bleu\_2 & Bleu\_3 & Bleu\_4 & ROUGE\_L & CIDEr & Match & Final Score\\
\midrule
ImagineLab*  & 100\%  & 0\%   & 0.76 & 75.00 & 0.75 & 0.69 & 0.63 & 0.57 & 0.71 & 0.12 & 34.35 &  0.6119 \\
LDM &  100\%  & 0\%  & \textbf{0.85} & \textbf{76.12} & \textbf{0.76} & \textbf{0.70} & \textbf{0.64} & \textbf{0.59} & \textbf{0.71} & \textbf{0.14} & \textbf{47.57} & \textbf{0.6638} \\ 
\hline
ImagineLab$^{*}$ & 5\%  & 0\% & 0.61 & 73.42 & 0.20 & 0.15 & 0.12 & 0.09 & 0.21 & 0.00 & 23.95 & 0.4864 \\
$\text{LDM}_0$ & 5\% & 0\% & 0.69 & 72.56 & 0.16 & 0.12 & 0.09 & 0.07 & 0.25 & 0.01 & 34.73 & 0.5231 \\

$\text{LDM}_1$ & 5\% & 20\%  & 0.75 & 72.86 & 0.69 & 0.62 & 0.57 & 0.52 & 0.70 & 0.08 & 36.61 & 0.6018 \\
$\text{LDM}_2$ & 5\% & 95\%  & \textbf{0.79} & \textbf{73.70} & \textbf{0.71} & \textbf{0.64} & \textbf{0.59} & \textbf{0.53} & \textbf{0.70} & \textbf{0.11} & \textbf{36.86} & \textbf{0.6157}\tiny(+9.26\%) \\
\bottomrule
\end{tabular}}
\label{tab:main_val}
\vspace{-3mm}
\end{table*}

\subsection{Self-Consistency Refinement}
\label{sec:self_consistency}
The SCR aims to refine the generated VQAs by providing a reliable score or filtering the unreliable QA pairs. Inspired by the intuition that with more hints about an object, the LDM could better predict the state of objects, the reliable score is computed via the distance between the original and the re-asked answers, which injects multiple hints retrieved from the scene graph in the prompt.

Specifically, we define the hints from the neighbors for any question corresponding to the $j$-th attribution $\mathcal{A}^j_i$ of object $o_i$, the hints are defined as the rest attributions and edges:
\begin{equation}
\label{eq:hints}
\begin{aligned}
    \mathcal{H}_{a} &= \{\mathcal{A}^{0}_i, \cdots \mathcal{A}^{j-1}_{i}, \mathcal{A}^{j+1}_{i}, \cdots \}, \\
    \mathcal{H}_{e} &= \{ \mathcal{E}_{i, 1}, \mathcal{E}_{i,2}, \cdots, \mathcal{E}_{i,n} \}, \\
    \mathcal{H} &= \{\mathcal{H}_{a}, \mathcal{H}_{e} \},
\end{aligned}
\end{equation}
where $\mathcal{H}_{a}$ are the hints about the node attributions and $\mathcal{H}_{e}$ indicates the hints of the edges. $\{\mathcal{H}_{a}, \mathcal{H}_{e}\}$ means the union of the hints. To avoid information redundancy and shorten the length of the prompt, we randomly select $k=4$ hints and inject to the prompt. Here is an example of the prompt with hints: \textit{``Consider the object $o_i$ is a truck, there is a car in the front stopped, there is a red light in the front, what's the moving status of the object $o_i$?”}. 
Then, we could compute the reliable score as follows:
\begin{align}
\label{eq:score}
    s = dist(\mathcal{M}(v, q, \pi_{\theta}), \mathcal{M}(v, \{q, \mathcal{H}\}, \pi_{\theta}) ),
\end{align}
In practice, for the distance function $dist$, we obtain the average sentence embedding via the fine-tuned LDM and define the cosine similarity as the distances. 
We also introduce the filter method, which drops the unreliable samples~(e.g. we manually set $s\leq 0.8$) and treats the reliable ones equally.
% \begin{equation}
% \label{eq:filter}
% s=\left\{
% \begin{aligned}
% 1,\quad s > \tau, \\
% 0,\quad s\leq \tau. \\
% \end{aligned}
% \right.
% \end{equation}
% Here we manually set the threshold $\tau=0.8$.

\section{Experiment}
\label{sec:exp}

\subsection{Benchmark and Evaluation Metric}
\textbf{DriveLM}~\cite{sima2023drivelm} is a large-scale language-based VQA benchmark for autonomous driving. We conduct experiments based on the DriveLM-nuScenes. The training set consists of 378K VQA pairs from 4072 scenes extracted from 696 video clips. The test set consists of 15k VQA pairs from 799 scenes extracted from 149 video clips from the nuScenes~\cite{caesar2020nuscenes} dataset. 
% The questions are about perception, prediction, planning, and behavior.
The evaluation metrics include accuracy to statistics the precision of the multiple-choice-question, the chatGPT evaluates the similarity of the prediction sentences of the planning task, the language score such as Bleu, ROUGE\_L, and CIDEr illustrated the prediction of the perception task, and the match score denotes the recall of the prediction objects. The final score is the weighted average of the above.

\subsection{Implementation Detail}
\label{sec:imp}
To demonstrate the effectiveness of our proposed semi-supervised training pipeline, we supervised train the LDM with 5\% (29 clips) annotations named $\text{LDM}_0$. Since our proposal could iteratively utilize the unlabeled data, we increment 20\%~(118 clips) and 75\%~(445 clips) unlabeled data to boost the $\text{LDM}_0$, named $\text{LDM}_1$ and $\text{LDM}_2$ respectively.
We also split a validation set including 4432 VQA pairs from 105 clips for easier verification. All the video clips are randomly selected from the DriveLM train split.
For the model training, we utilize the pre-trained InternVL2~\cite{chen2024internvl} as the foundation model and fine-tune the whole parameter except the visual encoder. All the experiments are trained on 16 Tesla A100 with batch size 1024, the learning rate and weight decay are set as 2e-5 and 5e-2 respectively.

\begin{figure*}[htbp]
    \centering
    \includegraphics[width=0.95\linewidth]{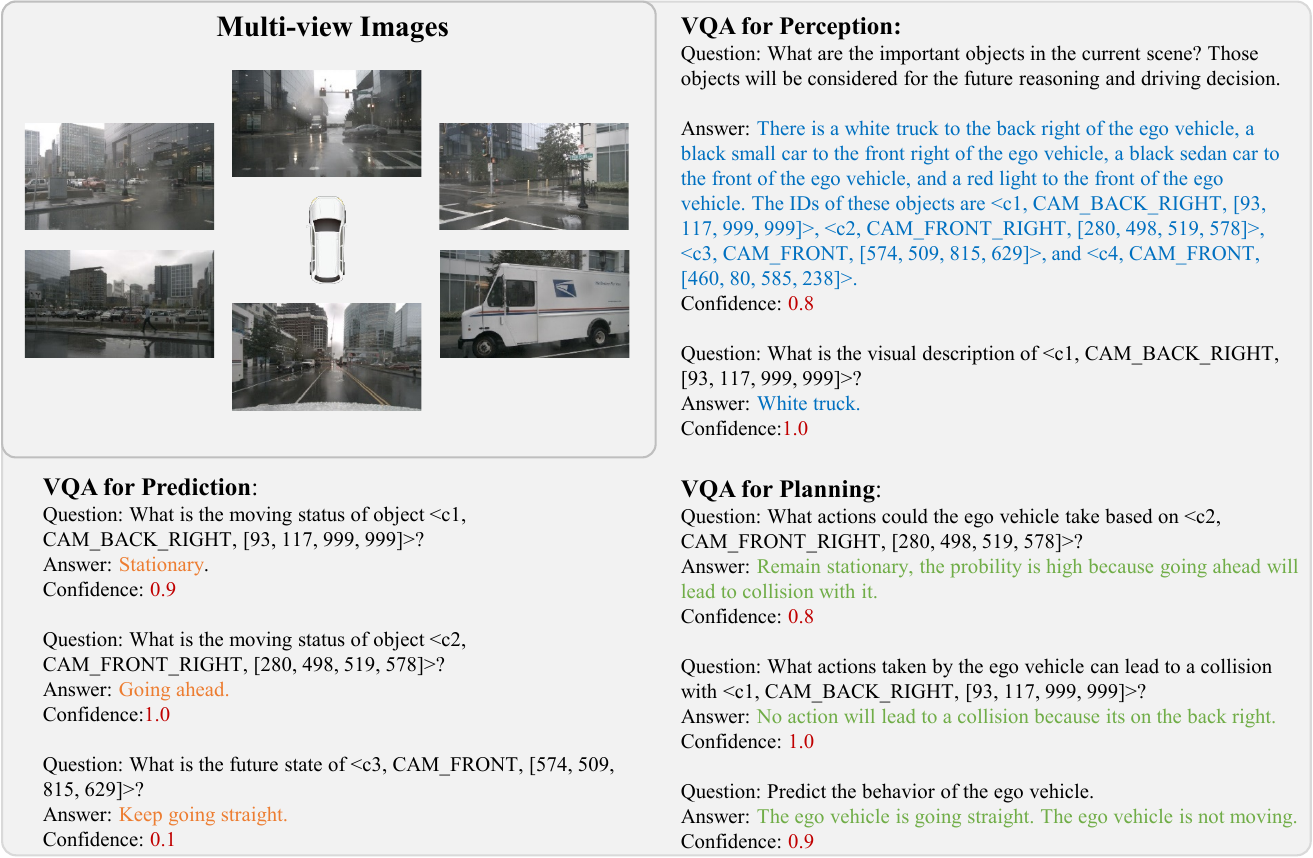}
    \vspace{-1mm}
    \caption{Visualization of the generated QA pairs on the unlabeled images, including perception/prediction questions to extract the node's attributions and planning questions for the edges. Each VQA pair contains a score obtained via proposed SCR.}
    \vspace{-4mm}
    \label{fig:vis}
\end{figure*}

\subsection{Main Result}
To demonstrate the effectiveness of our proposed LDM model and iterative training pipeline, we conduct experiments based on the full annotations and a few~(5\%) annotation data. The performance on the test set and validation set are shown in Tab.~\ref{tab:main_test} and Tab.~\ref{tab:main_val} respectively.

As shown in Tab.~\ref{tab:main_test}, we first display the supervised training results of our proposed LDM, to compare with the previous state-of-the-art methods, such as ImagineLab~\cite{li2024driving}, and MTMM~\cite{anonymous2024mmad}
We could observe that our proposal gains comparable performances against them and ranks second in the leaderboard\footnote{\url{https://huggingface.co/spaces/AGC2024/driving-with-language-official}}~(until the paper submitted data Sept. 15, 2024). Benefiting from the object selection mechanism mentioned in Sec.~\ref{sec:question}, we could obtain a much higher match score. 

Meanwhile, to demonstrate the effectiveness of our proposed iterative training pipeline in unlocking the potential of unlabeled data, we conduct experiments based on a few annotated data. Here the $\text{LDM}_0$, $\text{LDM}_1$, and $\text{LDM}_2$ indicate the results of each iteration as mentioned in Sec.~\ref{sec:imp}. Specifically, we first train the $\text{LDM}_0$ on the 5\% labeled data in a supervised manner, and then incrementally utilized 20\% and 75\% unlabeled data to train the $\text{LDM}_1$ and $\text{LDM}_2$ respectively via our proposed iterative training pipeline.
As shown in Tab.~\ref{tab:main_test}, we could observe that our proposal could continually improve the performance by using increasing unlabeled data. The $\text{LDM}_2$ achieves a 54.27\% final score and gains +9.42\% promotion without introducing extra annotations. Simultaneously, by utilizing the unlabeled data, our proposal gains comparable performances with 5\% annotations compared with the supervised training on 100\% annotations (e.g. 54.27\% versus 60.68\%).
Besides, we also conduct experiments on the validation set as shown in Tab.~\ref{tab:main_val}. The +9.26\% improvements further demonstrate the effectiveness of utilizing the unlabeled data.

\subsection{Ablation Study}
To analyze the impact of our iterative training pipeline and self-consistency refinement, we conduct ablation studies about the refinement and hints generation method.

First, we compare our SCR methods including scoring in E.q.~\ref{eq:score} and filter method, which obtain the reliable scores and drop the unreliable samples respectively. As shown in Tab.~\ref{tab:aba_score}, we could observe that compared with directly utilizing the predicted VQAs without post-process, our proposed SCR could gain improvements by a large margin~(e.g. 60.18\% versus 54.76\%), benefitting from the quality improvement of the generated VQAs. Besides, scoring surpasses the filter method because prediction with hints in the prompt may still obtain a wrong response, the filter method would drop the correct prediction under the situation.

Then, we conduct experiments based on variant hints generation in E.q.~\ref{eq:hints}, pruning the attributions and edges respectively as shown in Tab.~\ref{tab:aba_hints}. We could observe that with hints from attributions as well as edges together, the final score is preferable then the experiments with the attributions or edges individually. With more information in the prompt, including the object status and interaction, the model could obtain a better response. This further verifies our intuition that additional relevant hints could boost the model for a better understanding of objects.

\begin{table}[h!]
\centering
\caption{Performance comparison on the validation set with self-consistency refinement methods. "None" indicates without refinement.}
\vspace{-1mm}
\resizebox{\linewidth}{!}{
\begin{tabular}
{C{0.5cm}
C{0.3cm}
C{0.25cm}
C{0.3cm}
C{0.25cm}
C{0.25cm}
C{0.25cm}
C{0.25cm}
C{0.25cm}
C{0.25cm}
C{0.3cm}
C{0.5cm}
}
\toprule
Method & SCR & Acc. & C.G. & B1 & B2 & B3 & B4 & R.L. & CI. & Mat. & Score \\
\midrule
$\text{LDM}_0$ & ~~~- &  0.69 & 72.56 & 0.16 & 0.12 & 0.09 & 0.07 & 0.25 & 0.01 & 34.73 & 0.5231\\
$\text{LDM}_1$ & None & 0.72 & 60.31 & 0.72 & 0.66 & 0.60 & 0.54 & 0.70 & 0.08 & 35.88 & 0.5476 \\
$\text{LDM}_1$ & Filter & \textbf{0.77} & 72.21 & \textbf{0.71} & \textbf{0.65} & \textbf{0.59} & \textbf{0.54} & \textbf{0.70} & 0.08 & 33.02 & 0.5978 \\
$\text{LDM}_1$ & Score & 0.75 & \textbf{72.86} & 0.69 & 0.62 & 0.57 & 0.52 & \textbf{0.70} & 0.08 & \textbf{36.61} & \textbf{0.6018} \\
\bottomrule
\end{tabular}}
\label{tab:aba_score}
\vspace{-3mm}
\end{table}

\begin{table}[h!]
\centering
\caption{Performance comparison on the validation set about hints generation method. A. and E. are short of the attributions and edges respectively.}
\vspace{-1mm}
\resizebox{1.0\linewidth}{!}{
\begin{tabular}{C{0.5cm}p{0.3cm}<{\centering}C{0.25cm}C{0.45cm}C{0.25cm}C{0.25cm}C{0.25cm}C{0.25cm}C{0.25cm}C{0.25cm}C{0.35cm}C{0.5cm}}
\toprule
Method & Hints & Acc. & C.G. & B1 & B2 & B3 & B4 & R.L. & CI. & Mat. & Score \\
\midrule
$\text{LDM}_0$ & ~~~- &  0.69 & 72.56 & 0.16 & 0.12 & 0.09 & 0.07 & 0.25 & 0.01 & 0.35 & 0.5231 \\
$\text{LDM}_1$ & None & 0.72 & 60.31 & 0.70 & 0.64 & 0.58 & 0.52 & 0.69 & 0.07 & 35.88 & 0.5441 \\
$\text{LDM}_1$ & ~~~A. & 0.73 & 63.07 & \textbf{0.71} & \textbf{0.65} &\textbf{0.59} & \textbf{0.53} & \textbf{0.70} & 0.07 & 38.34 & 0.5585 \\

$\text{LDM}_1$ & ~~~E. & 0.73 & 63.03 & 0.69 & 0.63 & 0.58 & 0.52 & 0.69 & \textbf{0.08} & \textbf{38.37} & 0.5602 \\

$\text{LDM}_1$ & A.+E. & \textbf{0.75} & \textbf{72.86} & 0.69 & 0.62 & 0.57 & 0.52 &\textbf{0.70} & \textbf{0.08} & 36.61 & \textbf{0.6018} \\
\bottomrule
\end{tabular}}
\label{tab:aba_hints}
\end{table}

\subsection{Visualization}
We visualize a sample with generated predictions in Fig.~\ref{fig:vis}. We could observe that in the generated QA pairs, most of them are reliable with high scores, but still contain incorrect predictions for a confusing scenario. For example, the truck in the front is moving forward, and the model would predict its moving status as \textit{"going ahead."}. With hints in the prompt, the model knows there is a red light and crossing car in the front, and the prediction will be \textit{"turning right."}. The SCR could score this confusion sample a lower weight.

\section{Conclusion}
In this paper, we propose improving the Language Driving Model~(LDM) by unlocking the value of large-scale unlabeled data. 
Specifically, we propose an iterative training pipeline to train the LDM by iteratively utilizing incremental unlabeled data in a semi-supervised manner, in which we propose a series of template-based prompts to generate diversity and valuable VQAs. To refine the generated pseudo label, we propose Self-Consistency Refinement~(SCR) via retrieving the graph-based hints. Extensive experiments and huge promotions demonstrate its effectiveness. We hope our proposal will inspire future research in scaling up the training data without relying upon massive human annotations.

\section*{ACKNOWLEDGMENT}
The work is partially supported by Shenzhen Science and Technology Program JCYJ20220818103001002, the Guangdong Key Laboratory of Big Data Analysis and Processing, Sun Yat-sen University, China, and by the High-performance Computing Public Platform (Shenzhen Campus) of Sun Yat-sen University.

\bibliographystyle{IEEEtran}
\bibliography{ref}

\end{document}